\documentclass[12pt,journal,compsoc,onecolumn]{IEEEtran}

\usepackage{lineno,hyperref}
\usepackage{tikz}
\usetikzlibrary{calc}
\usepackage{relsize}
\usepackage[noadjust]{cite}

\tikzset{fontscale/.style = {font=\relsize{#1}}
    }

\modulolinenumbers[5]
\usepackage{graphicx}
\usepackage{amsmath}
\usepackage{amssymb}
\usepackage{color}
\newcommand{\cmark}{\ding{51}} 
\newcommand{\xmark}{\ding{55}} 
\usepackage{pifont}
\let\OLDthebibliography\thebibliography
\renewcommand\thebibliography[1]{
  \OLDthebibliography{#1}
  \setlength{\parskip}{1.5pt}
  \setlength{\itemsep}{0pt plus 0.3ex}
}
\begin{document}

\def\x{{\mathbf x}}
\def\L{{\cal L}}

\def\S{{\cal S}}
\def\V{{\cal V}}
\def\N{{\cal N}}
\def\T{{\cal T}}
\def\betaa{{\bf \beta}}
\def\w{{\bf w}}                 
                  \def\K{{\cal K}}

\def\betaaa{{\hat{\beta}}}

\title{Action Recognition with Deep Multiple Aggregation Networks} 
\author{Ahmed Mazari  \ \ \ \ \ \ \ \ \  \ \ \ Hichem Sahbi\\
Sorbonne University, CNRS, LIP6\\
F-75005, Paris, France 
}

\maketitle

\begin{abstract}
 Most of the current action recognition algorithms are based on deep networks which stack multiple convolutional, pooling and fully connected layers. While convolutional and fully connected operations have been widely studied in the literature, the design of pooling operations that handle action recognition, with different sources of temporal granularity in action categories, has comparatively received less attention, and existing solutions rely mainly on max or averaging operations. The latter are clearly powerless to fully exhibit the actual temporal granularity of action categories and thereby constitute a bottleneck in classification performances. \\
 \indent In this paper, we introduce a novel hierarchical pooling design that captures different levels of temporal granularity in action recognition. Our design principle is coarse-to-fine and achieved using a tree-structured network; as we traverse this network top-down, pooling operations are getting less invariant but timely more resolute and well localized. Learning the combination of operations in this network --- which best fits a given ground-truth --- is obtained by solving a constrained minimization problem whose solution corresponds to the distribution of weights that capture the contribution of each level (and thereby temporal granularity) in the global hierarchical pooling process. Besides being principled and well grounded, the proposed hierarchical pooling is also video-length and resolution agnostic. Extensive experiments conducted on the challenging  UCF-101, HMDB-51 and JHMDB-21 databases corroborate all these statements.
\end{abstract}

\hspace{1cm}{\small {\bf Keywords---} Multiple aggregation design, 2-stream networks, action recognition}
  
 \section{Introduction}
 
Action recognition is standing as one of the most challenging problems in video processing which consists in assigning one or multiple semantic categories to moving objects. This task is difficult as scenes are acquired under extremely challenging conditions including cluttered backgrounds, viewpoint changes, illumination variations, poor camera sensor quality and resolution, and this affects the accuracy of multiple related applications such as scene understanding~\cite{scene01,scene02,sahbijstars17,scene03,icip2001}, video surveillance~\cite{video_surv01,sahbiicip09,video_surv02,sahbiigarss12b,sahbiigarss12}, video caption generation and retrieval~\cite{sahbiicpr18,video_capt02,sahbispie2004,video_capt03,sahbiclef08,video_capt04,sahbiijmir15,video_capt05,sahbikpca06,vid_retri01,sahbijmlr06,vid_retri02,sahbiaccv2010,sahbiclef13,vid_retri04,sahbisc7,vid_retri05,sahbicassp11,fuzzy05} as well as human computer interaction and robotics~\cite{vid_robotics01,vid_robotics02,vid_robotics03,vid_robotics04}. Most of the existing action recognition solutions are based on machine learning (ML)~\cite{temporalpyramid_detec,mkl_action,temporal_pyramid,lingsahbi2013,superived_dic_action,multi_svm,sahbicvpr08a,sahbiphd,sahbipr19,sahbifleuret02}; their general recipe consists in learning functions that map visual content representations of frame sequences (either handcrafted or learned~\cite{hog,of,bag_features,Fisher}) into categories using widely used ML algorithms such as random forests, support vector machines and deep networks \cite{lingsahbi2013,Resnet16,phong,rbf,polynomial_kernel,temporalpyramid_detec,tnnls19,imagenet,Mict,action_localization02,action_localization03,action_localization05,action_localization01,icassp2017b,lingsahbieccv2014,lingsahbiicip2014,speech_reco01,sahbipr2012,speech_reco02,image_class01,icml08,sahbiiccv17}. \\ 

Among the ML solutions --- for action recognition  --- those based on deep networks are currently witnessing a major interest \cite{Resnet16,temporalpyramid,temporalpyramid_detec,speech_reco01,speech_reco02,image_class01,image_class02} but their success is tributary to the availability of large amount of labeled training data and also the appropriate choice of their architectures including convolutional and recurrent ones \cite{kin3d,pose,spresnet16,spresnetmulti17,temporalpyramid}. In particular, convolutional networks are designed by stacking multiple convolutional, pooling and fully connected layers; successful architectures for action recognition include two-stream 2D/3D convolutional neural networks (CNNs) operating on appearance and motion flows, and CNNs combined with Long Short-Term Memory (LSTM) networks~\cite{ullah2017} that capture coarse temporal structure of actions as well as 3D CNNs~\cite{kin3d} which capture fine (local) temporal structures. However, and beside issues  related to scarcity of labeled data and the large number of training parameters (especially in 3D CNN models), the effort in the design of deep networks, that capture the relevant motion information in videos, has been focused essentially on optimizing their convolutional and fully connected layers\footnote{Convolutions and multi-layer perceptron have been largely studied since the early age of artificial neural networks and also in other problems in image processing including wavelet filter design.} while {\it  comparatively} the design of optimized pooling layers received less attention especially on non-vectorial data including video sequences. The difficulty in designing architectures with suitable pooling (a.k.a aggregation) operators, particularly on video sequences stems from the eclectic properties of videos (namely their duration, temporal resolution and velocity of moving objects as well as the granularity of their action categories) and this makes pooling design very challenging. This challenge is further exacerbated by the lack of labeled video data (covering all the variability) compared to other neighboring problems such as image classification that benefit from labeled sets which are at least an order of magnitude larger compared to the current action recognition datasets while the task is intrinsically far more challenging; as a result, these action recognition models are more subject to overfitting. \\

\begin{figure}[h!]
    \centering
    \includegraphics[width=0.5\linewidth]{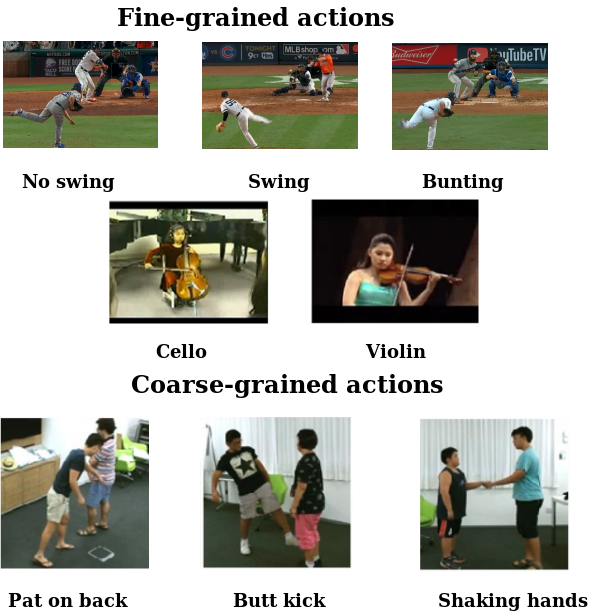}
    \caption{{\it Examples of {\it fine} and {\it coarse-grained} actions. The first row shows three action categories from the MLB-YouTube dataset \cite{MLB-YouTube}: ``No swing'', ``Swing'' and ``Bunting'' which are difficult to distinguish as they have very small differences. The second row shows two instrument playing actions from the UCF-101 dataset \cite{ucf}: ``cello'' and ``violin'' which are also difficult to distinguish as their arm/hand locations and directions are similar. In contrast, the third row shows ``Pat on back'', ``Butt kick'' and ``Shaking hand'' actions (taken from NTU RGB+D dataset \cite{ntu}) which are relatively easier to distinguish.}}
    \label{fig:fine_coarse}
  \end{figure}

\indent In order to attenuate such effect, pooling methods \cite{pooling01,pooling02,pooling03} have been designed, and most of them are based on global  measures including max and averaging operators. Pooling plays a key role in reducing the dimensionality of convolutional feature maps and thereby the number of training parameters and enhances the resilience, of the learned CNN representations, to the lack of training data and to the acquisition conditions. However, it comes at the detriment of some relative loss in the discrimination power especially when video data belong to fine-grained action categories. Indeed, pooling contributes in diluting (averaging) convolutional features which are highly important in discriminating fine-grained categories and these averaging operators are rather more appropriate for coarse-grained actions (see Fig. \ref{fig:fine_coarse}). Alternative and more recent solutions \cite{kin3d,spresnet16,spresnetmulti17} rely on sampling and stacking CNN features in order to build {\it spectrogram-like} fixed length representations that also preserve the granularity of video actions. Nonetheless, both methods suffer from several drawbacks; on the one hand, pooling methods based on global statistical measures are time/duration agnostic (and hence invariant) but less discriminating while spectrogram-like methods are discriminating but time/duration aware (less invariant) and highly sensitive to the aforementioned video acquisition conditions and may result into a loss of informations, especially when videos are not well resolute. \\ 
 
\indent A more suitable pooling should gather the advantages of these two families of methods while discarding their inconvenients. Following this goal, we consider in our work  a hierarchical aggregation scheme that describes moving scenes at multiple temporal granularities while also being resilient to their highly variable acquisition conditions. Top levels in this hierarchical aggregation provide orderless (invariant) but less discriminating motion and appearance representations which capture coarse-grained action categories (as global averaging techniques~\cite{temporalpyramid})  {\it while} bottom levels correspond to fine-grained, timely resolute  and order-sensitive  video representations (as spectrogram based techniques~\cite{temporalpyramid}). The design principle of our proposed solution  is {\it coarse-to-fine} and allows us to capture a gradual change of invariance and granularity; as we traverse the hierarchy top-down, our video representations are getting less invariant but timely more resolute and fine-grained. However, knowing a priori which levels in this hierarchy are the most appropriate in order to capture the actual granularity of our video data  is challenging  and also {\it combinatorial}; hence, learning this combination ``end-to-end'' and in a differentiable manner is rather more appropriate. \\

\indent Considering this line of research, other related works \cite{STP_CV17,tp_segment,stp_cnn,tp_concat,tp_scale} try to model granularity of actions in videos by incorporating specific modules into CNNs. The method in \cite{STP_CV17} samples, from each video, frames as well as their associated optical flow components and adds a spatio-temporal  module to CNN in order to capture hierarchical relationships between appearance and motion features. The method in \cite{tp_segment} stacks a hierarchical temporal pooling layer on the top of motion and appearance CNN streams in order to build fixed-length video representations. In \cite{stp_cnn}, authors sample a set of frames by first splitting videos into segments and taking frames from each segment, and build a spatial pyramid to extract multi-scale appearance features from different convolutional layers. These features are then concatenated and fed to a three level temporal hierarchy. The work in \cite{tp_concat} samples video frames at different temporal resolutions, and feeds them to a 3D CNN in order to extract their respective features followed by a temporal hierarchy which down-samples and concatenates the resulting features. Finally, the method in \cite{tp_scale} achieves frame sampling followed by a temporal pooling in order to build features at different pyramidal levels; the resulting features are afterwards fed to a temporal relational layer that  groups these features at different scales. While all these methods rely on a hierarchical temporal  aggregation scheme, none of them considers the issue of learning the best combination of levels in these temporal aggregation hierarchies, and this turns out to be highly effective as shown in the following sections.\\

\indent In this paper, we introduce a novel scheme for action recognition based on Deep Multiple Aggregation Networks.  Given a hierarchy of aggregation operations,  the goal is to learn a combination of these operations that best fits a given action recognition ground-truth.  We solve this problem by minimizing a constrained objective function whose parameters correspond to the distribution of weights through multiple aggregation levels; each weight captures the granularity of its level and its contribution in the global learned video representation.  Besides handling aggregation at different levels, the particularity of our solution resides in its ability to handle variable length videos (without any up or down-sampling) and thereby makes it possible to fully benefit from the whole frames in videos. \\

The rest of this paper is organized as follows. First, we describe in Section~2 our motion and appearance streams used to build frame-wise representations. Then, we introduce in Section~3 our main contributions; two methods  --- based on linear/nonlinear kernel combination and ``end-to-end'' two stream CNN training --- that aggregate and combine the obtained frame-level representations into temporal pyramids in order to achieve action recognition. Finally, we show the validity of these contributions through extensive experiments using standard and challenging video datasets including UCF-101, HMDB-51 and JHMDB-21. 

 \begin{figure*}[hbp!]
    \centering
    \hspace{-0.15cm}     \includegraphics[angle=90,width=0.35\linewidth]{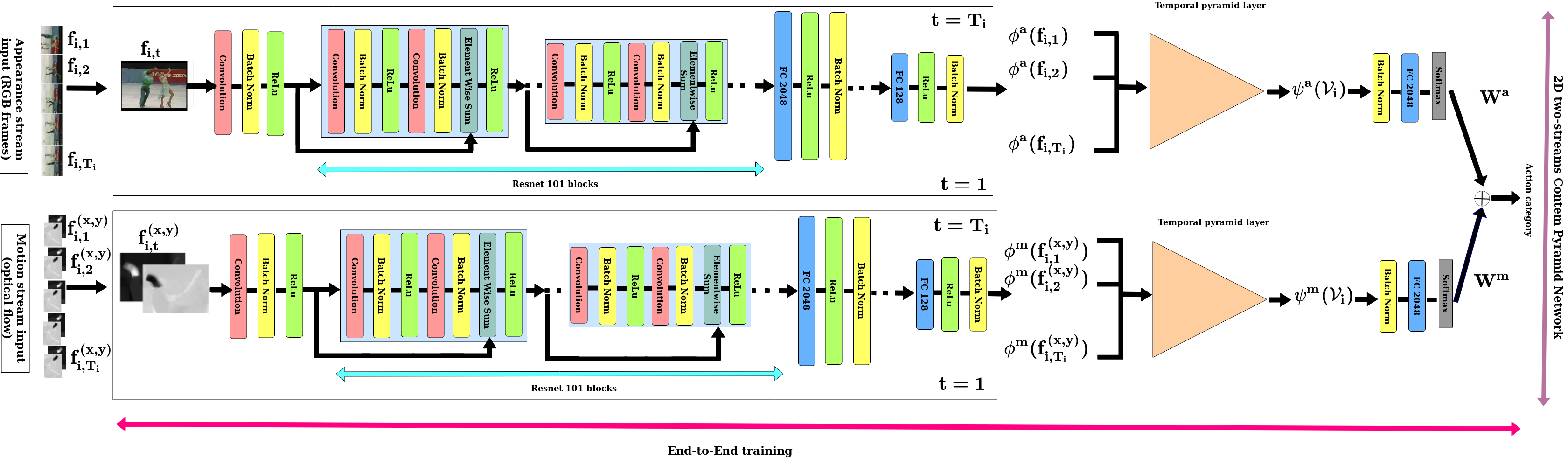}
    \caption{\it Our two stream network including a ResNet block, a temporal pyramid block and ``batch norm+fully connected+softmax+late fusion'' layers. The temporal pyramid block achieves pooling either by weighted averaging or weighted concatenation (see Eq.~\ref{eq1} and also Fig~\ref{fig:concat_vs_averaging}) ({\bf Better to zoom the PDF version}).}
    \label{fig01}
  \end{figure*}

  \begin{figure}
    \centering
    \includegraphics[angle=-90,width=0.2\linewidth]{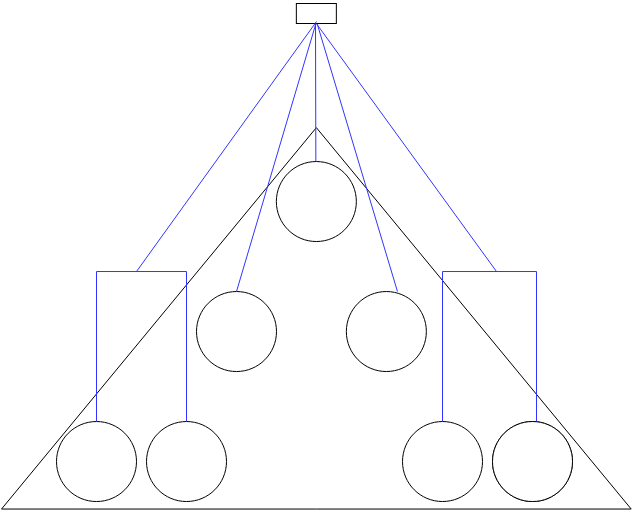}\hspace{1cm} \includegraphics[angle=-90,width=0.2\linewidth]{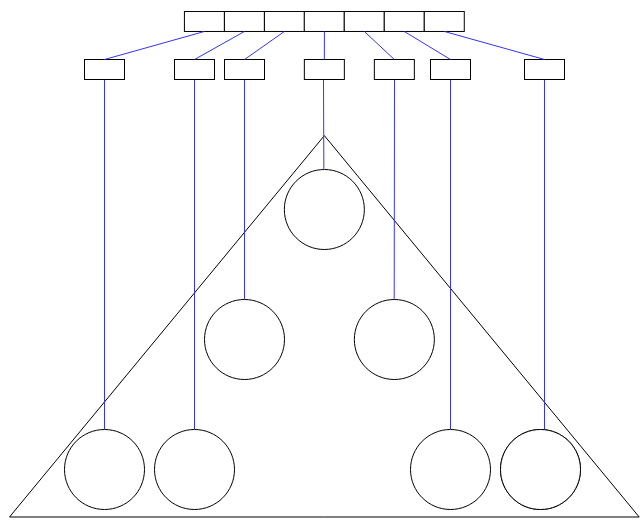}
    \caption{\it  Aggregation by ``averaging'' vs.  aggregation by ``concatenation''.}
    \label{fig:concat_vs_averaging}
\end{figure}

\section{Frame-wise Two-Stream Video description at a Glance}
We consider a collection of videos $\S = \{\V_i \}_{i=1}^n$ with each one being a sequence of frames $\V_i=\{f_{i,t}\}_{t=1}^{T_i}$  and a set of action categories (a.k.a classes or categories) denoted as ${\cal C} = \{1,\dots, C\}$. In order to describe the visual content of a given video $\V_i$, we rely on a two-stream process (see Fig.~\ref{fig01}); the latter provides a complete description of appearance and motion that characterizes the spatio-temporal aspects of moving objects and their interactions. The output of the appearance stream (denoted as $\{\phi_a(f_{i,t})\}_{t=1}^{T_i} \subset \mathbb{R}^{2048}$) is based on the deep residual network (ResNet-101) trained on ImageNet \cite{imagenet} and fine-tuned on UCF-101 \cite{ucf} while the output of the motion stream (denoted as $\{\phi_m(f_{i,t})\}_{t=1}^{T_i} \subset \mathbb{R}^{2048}$) is also based on the ResNet 101 network but trained on optical flow image pairs \cite{pretrained_resnet101_ucf,segment_net}; these pairs correspond to the horizontal and the vertical displacement fields which are linearly transformed in order to make their ranges between 0 and 255.\\

\indent Following the line in \cite{segment_net} and in order to adapt the pretrained ResNet-101 to optical flow data, we slightly update the input layer of the original ResNet\footnote{Already available/pretrained on ImageNet to capture the appearance.}. Indeed, the number of channels is reset to $20$ instead of $3$ in the original ResNet; the initial weights of these 20 channels are obtained by averaging the 3 original (appearance) channel weights and by replicating their values through the 20 new motion channels. During training, closely related methods (namely \cite{segment_net}) split each video into $N$ continuous segments, and for each segment, a frame $f$ is randomly selected to feed an appearance stream ResNet and a stack of optical flow is also taken (starting from $f$) as an input to the motion stream.  
In the setting of \cite{segment_net}, scores obtained from the softmax layers of motion and appearance streams are fused through different frames using a {\it segmental consensus function} in order to make class prediction at the video level; in other words, for each test video, 19 frames\footnote{The reason for choosing 19 frames is explained by the fact that the minimum number of video frames in UCF-101 is 28, hence 19 is the maximum number from which a stack of 10 optical flow frames can be taken.} are uniformly sampled and passed through appearance and motion streams and their  scores are combined as votes for all the action categories. As shown subsequently, and in contrast to \cite{segment_net}, our proposed method relies on a different aggregation scheme that models {\it coarse as well as fine grained action categories}; besides, our method does not require any frame (re)sampling  -- which may degrade performances (as also shown later in experiments) -- indeed, our method effectively leverages the entire set of video frames.

\def\N{{\cal N}}
\def\T{{\cal T}}
\def\betaa{{\bf \beta}}
\def\w{{\bf w}}                 

  \begin{figure}[h!]
  \begin{center}
  \begin{tikzpicture}[font=\footnotesize][!h]
\centering
   \node at (0,0) {\includegraphics[width=0.7\columnwidth]{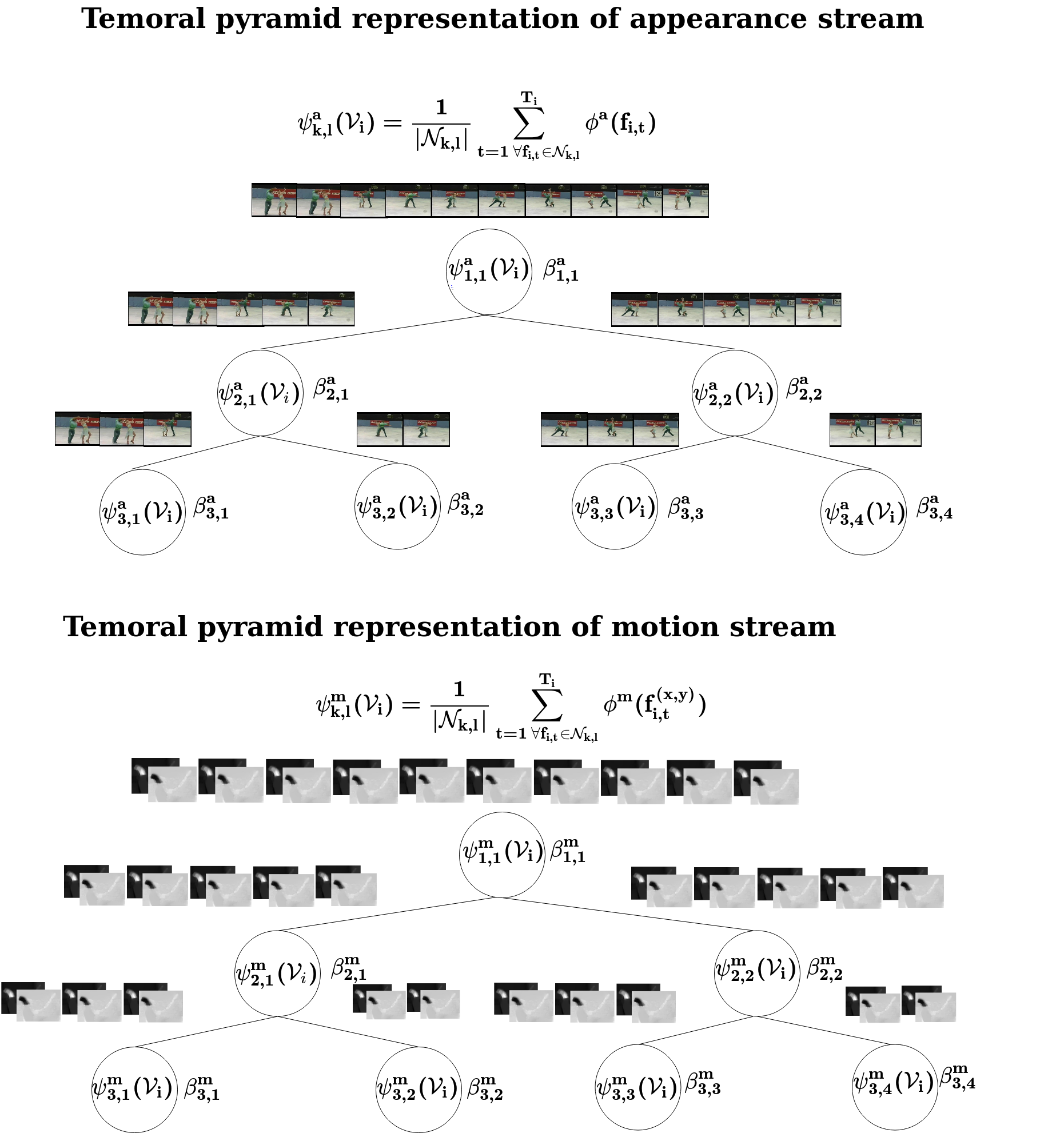}};
   \draw [color=white,fill=white,xshift=-2cm,opacity=1] (-3.5,4.8)--(7,4.8)--(7,7.1)--(-3.5,7.1)--cycle;
  \node[draw,color=white,text width=6.2cm,text=black] at (0,6.0) {Temporal pyramid  (appearance stream) };
    \node[draw,color=white,text width=5cm,text=black] at (0,5.2) {$\displaystyle \psi^a_{k,l}(\V_i)=\frac{1}{|\N_{k,l}|}\sum_{t \in \N_{k,l}} \phi_a(f_{i,t})$};
   \draw [color=white,fill=white,xshift=-2cm,opacity=1] (-3.7,-2.2)--(7,-2.2)--(7,0.0)--(-3.7,0.0)--cycle;
   \node[draw,color=white,text width=5.5cm,text=black] at (0,-0.8) {Temporal pyramid  (motion stream) };
       \node[draw,color=white,text width=5cm,text=black] at (0,-1.6) {$\displaystyle \psi^m_{k,l}(\V_i)=\frac{1}{|\N_{k,l}|}\sum_{t \in \N_{k,l}} \phi_m(f_{i,t})$};
  \end{tikzpicture}
\end{center} 
     \caption{\it This figure shows frame aggregation at each node of the temporal pyramid for appearance (top) and motion streams (bottom).}\label{fig0001}
  \end{figure}
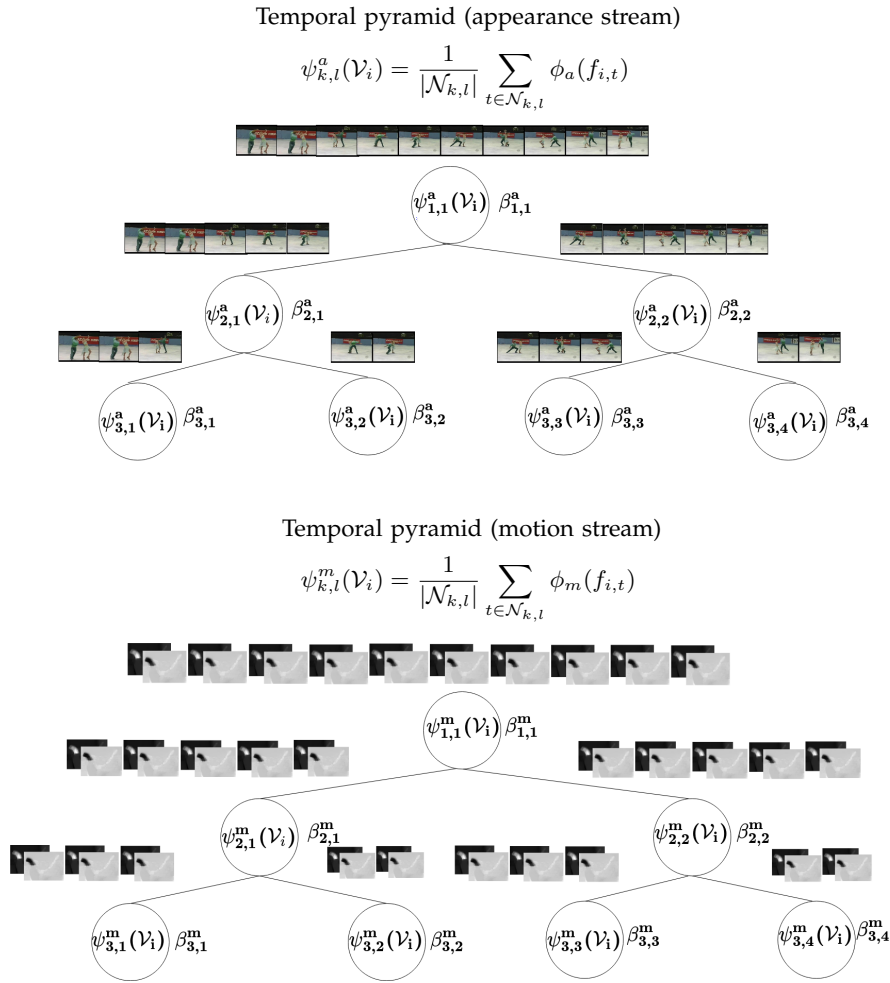

\section{Multiple Aggregation Learning}

\indent Given a video $\V_i$ with $T_i$ frames, we define $\N$ as a tree-structured network with depth up to $D$ levels and width up to $2^{D-1}$. Let $\N = \cup_{k,l} \N_{k,l}$ with $\N_{k,l}$ being the $k^{th}$ node of the $l^{th}$ level of $\N$; all nodes belonging to the $l^{th}$ level of $\N$ define a partition of the temporal domain $[0,T_i]$ into $2^{l-1}$ equally-sized subdomains. A given node $\N_{k,l}$ in this hierarchy aggregates the frames that belong to its underlying temporal interval. Each node $\N_{k,l}$ also defines an appearance and a motion representation respectively denoted as  $\psi^a_{k,l}(\V_i)$, $\psi^m_{k,l}(\V_i)$ and set as $\psi^a_{k,l}(\V_i)=\frac{1}{|\N_{k,l}|}\sum_{t \in \N_{k,l}} \phi_a(f_{i,t})$, $\psi^m_{k,l}(\V_i)=\frac{1}{|\N_{k,l}|}\sum_{t \in \N_{k,l}} \phi_m(f_{i,t})$; see Fig.~\ref{fig0001}. Depending on the level in $\N$, each representation captures a particular temporal granularity of motion and appearance into a given scene; it is clear that top-level representations capture coarse visual characteristics of actions while bottom-levels (including leaves) are dedicated to fine-grained and timely-resolute sub-actions. Knowing a priori which levels (and nodes in these levels) capture the best -- a given action category  -- is not trivial. In the remainder of this section, we introduce a novel learning framework which achieves multiple aggregation design and finds the best combination of levels and nodes in these levels that fits  different temporal granularities of action categories.\\
 
 \indent Considering the motion stream, we define -- for each node $\N_{k,l}$ -- a set of variables $\beta_m=\{\betaa_{k,l}^m\}_{k,l}$ (with $\betaa_{k,l}^m \in [0,1]$ and $\sum_{k,l} \betaa_{k,l}^m= 1$) which measure the importance (and hence the contribution) of $\psi^m_{k,l}(\V_i)$ in the global motion representation of $\V_i$ (denoted as $\psi^m(\V_i)$). Precisely, two variants are considered for $\psi^m$ 
\begin{equation}\label{eq1}
  \begin{array}{lll}
   \textrm{(*)}  &  \psi^m(\V_i) & = \displaystyle \left(\betaa_{1,1}^m \psi_{1,1}^m(\V_i)  \dots  \betaa_{k,l}^m \psi_{k,l}^m(\V_i) \dots \right)^\top \\
                 &  & \\
   \textrm{(**)}   & \psi^m(\V_i) &= \displaystyle \sum_{k,l} \betaa_{k,l}^m \psi^m_{k,l}(\V_i). 
                     \end{array}
                   \end{equation}                
                   \noindent As shown  above, the variant in (*) corresponds to a concatenation scheme while (**) corresponds to averaging.  Similarly to motion, we define the aggregations and the set of variables $\betaa_a=\{\betaa_{k,l}^a\}_{k,l}$  associated to appearance stream. In the remainder of this section, and unless explicitly mentioned, the symbols $m$, $a$ are omitted in the notation and all the subsequent formulation is applicable to motion as well as appearance streams.
                  \def\K{{\cal K}}
\subsection{Shallow multiple aggregation learning} 
In this section, we consider all the representations $\{\psi_{k,l}(.)\}_{k,l}$ fixed on all the video frames, and only the mixing parameters in $\betaa$ are allowed to vary.  Given the set of action categories ${\cal C}=\{1,\dots,C\}$;  we train multiple classifiers (denoted $\{g_c\}_{c \in {\cal C}}$) on top of these level-wise representations. In practice, we use maximum margin classifiers whose kernels correspond to  combinations of elementary kernels dedicated to $\{{\cal N}_{k,l}\}_{k,l}$. These classifiers are suitable choices as they allow us to weight the impact of nodes in the hierarchy $\cal N$ and put more emphasis on the most relevant granularity of the learned representations. Hence, depending on the granularity of action categories, these classifiers will prefer top or bottom levels of $\cal N$. \\

\indent Considering a training set of videos $\{(\V_i,y_{ic})\}_i$ associated to an action category c, with $y_{ic} =+1$ if $\V_i$ belongs to the category $c$ and $y_{ic} =-1$ otherwise, the max margin classifier associated to this action category $c$ is given by $g_c(\V) = \sum_{i} \alpha_i^c  y_{ic}  \K(\V,\V_i) + b_c$, here $b_c$ is a shift, $\{\alpha_i^c\}_i$ is a set of positive parameters and $\K$ is a positive semi-definite (p.s.d) kernel \cite{kernels2004}. In order to combine different nodes in the hierarchy $\cal N$ and hence design appropriate aggregation, we consider multiple representation learning that generalizes~\cite{MKL}  both to linear and nonlinear combinations (see also~\cite{sahbiicassp15,mklimage2017,sahbiicassp16b}). Its main idea consists in finding a kernel $\K$ as a combination of p.s.d elementary kernels $\{\kappa(.,.)\}$ associated to $\{\N_{k,l}\}_{k,l}$. Considering the two map variants in Eq.~(\ref{eq1}), we define these kernels as
 
\begin{equation} 
  \begin{array}{lll}   
    \K(\V,\V') &=&\displaystyle \sum_{l} \sum_{k}\beta_{k,l} \ \kappa(\psi_{k,l}(\V),\psi_{k,l}(\V')) \\
                  \K(\V,\V') &=&\displaystyle \sum_{l,l'} \sum_{k,k'}\beta_{k,l} \beta_{k',l'} \ \kappa(\psi_{k,l}(\V),\psi_{k',l'}(\V')). 

                   \end{array}\label{eq0000}
\end{equation}
As $\beta_{k,l} \in [0,1]$, the kernel $\K$ is p.s.d resulting from the closure of the p.s.d of $\kappa$ w.r.t the sum and the product. Using $\K$, we train the max margin classifiers $\{g_c\}_{c\in {\cal C}}$ whose kernels (in Eq.~\ref{eq0000}) correspond to level-wise linear (resp. cross-wise nonlinear) combinations of elementary kernels dedicated to $\{\N_{k,l}\}_{k,l}$. Hence, using a maximum margin formulation,  we find the parameters $\beta=\{\beta_{k,l}\}_{k,l}$ and $\{\alpha_i^c\}_{i,c}$ by solving 
\begin{equation} 
  \begin{array}{ll}
    \displaystyle 
    \min_{0 \leq \beta \leq 1, \|\beta \|_1=1,\{\alpha_i^c\}} &  \displaystyle  \frac{1}{2} \sum_c \sum_{i,j} \alpha_{i}^c \alpha_j^c y_{ic} y_{jc}  \K(\V_i,\V_j) - \sum_i \alpha_i^c \\
    \textrm{s.t.}              &   \displaystyle \alpha_{i}^c\geq 0,  \ \ \ \ \ \  \sum_{i} y_{ic} \alpha_{i}^c=0, \ \ \  \forall i, c. 
\end{array}\label{optimiz}
\end{equation}
As the problem  in Eq.~\ref{optimiz} is not convex w.r.t $\beta$, $\{\alpha_i^c\}$ taken jointly and convex when taken separately, an EM-like iterative optimization procedure can be used: first, parameters in $\beta$ are fixed and the above problem is solved w.r.t $\{\alpha_i^c\}$ using quadratic programming (QP), then  $\{\alpha_i^c\}$ are fixed and the resulting problem is solved w.r.t  $\beta$ using either linear programming for (*)  and QP for (**). This iterative process stops when the values of all these parameters remain unchanged or when it reaches a maximum number of iterations.  

\subsection{Deep multiple aggregation learning}
In this section, we consider an end-to-end framework that learns the parameters $\betaa_m$ and $\betaa_a$  together with (i) the ResNet parameters (denoted as $\alpha_a$, $\alpha_m$)\footnote{In the rest of this paper, the notation $\alpha$ refers to the ResNet parameters and not the max margin classifiers anymore.}, (ii) the MLP+softmax parameters (denoted as $\gamma_a$, $\gamma_m$)  as well as (iii) the mixing parameters (referred to as $\w_a$ and $\w_m$) which respectively capture the importance of appearance and motion streams in action recognition. Considering $E$ as the regularized cross-entropy loss\footnote{Regularization is achieved using $\ell_2$ weight decay.} associated to our complete network (in Fig. \ref{fig01}), we find the optimal $\alpha=\{\alpha_m,\alpha_a\}$, $\betaa=\{\betaa_m,\betaa_a\}$, $\gamma=\{\gamma_m,\gamma_a\}$ and $\w=\{\w_m,\w_a\}$ by solving the following constrained minimization problem
\begin{equation}\label{eq0}
\begin{array}{ll}
  \displaystyle   \min_{\alpha,\beta,\gamma,\w} &  E(\alpha,\beta,\gamma,\w)  \\
  \textrm{s.t.}  & 0 \leq \beta_{k,l}^m \leq 1,  \ \ \  \displaystyle \sum_{k,l} \beta^m_{k,l} = 1 \\
                 &  0 \leq \beta_{k,l}^a \leq 1,  \ \ \ \displaystyle \sum_{k,l} \beta^a_{k,l} = 1. 
 \end{array} 
\end{equation}
\def\betaaa{{\hat{\beta}}}
In spite of having many differences w.r.t usual losses used in deep learning, this objective function can still be solved using gradient descent and backpropagation. However, many differences exist and should be carefully tackled; indeed, whereas the forward step can be achieved, gradient backpropagation  (through our multiple aggregation layer) should be achieved while considering videos with a varying number of frames. Besides, constraints on $\beta's$ should also be handled. In what follows, we discuss all these updates in the optimization process. \\ 

\noindent {\bf Optimization.} Considering $\rho()$ as the output of the final layer of our deep network and  considering $\frac{\partial E}{\partial \rho}$ available, the gradients $\frac{\partial E}{\partial \w}$, $\frac{\partial E}{\partial \gamma}$ (w.r.t the preceding mixing and MLP layers) could easily be obtained using a straightforward application of the chain rule (as already available in the used PyTorsh tool). However,    $\frac{\partial E}{\partial \beta}$, $\frac{\partial E}{\partial \alpha}$ cannot be obtained straightforwardly; on the one hand, any step following the gradient $\frac{\partial E}{\partial \beta}$ should preserve equality and inequality constraints in Eq.~(\ref{eq0}) while a direct application of the chain rule provides us with a surrogate gradient which ignores these constraints. On the other hand, the variable number of frames for different training videos requires a careful update of $\frac{\partial E}{\partial \alpha}$ as shown subsequently.\\

\noindent {\bf Constraint implementation.} In order to implement the equality  and inequality constraints during the optimization of the objective function (\ref{eq0}), we consider a re-parametrization as $\beta_{k,l}^m= h(\betaaa_{k,l}^m)\slash {\sum_{k',l'} h(\betaaa_{k',l'}^m)}$ for some $\{\betaaa_{k,l}^m\}_{k,l}$ with $h$ being strictly monotonic real-valued (positive) function and this allows free settings of the parameters  $\{\betaaa_{k,l}^{m}\}_{k,l}$ during optimization while guaranteeing $\beta_{k,l}^{m} \in [0,1]$ and $\sum_{k,l}  \beta_{k,l}^{m}=1$. During back-propagation, the gradient of the loss $E$ (now w.r.t $\betaaa$'s) is updated using the chain rule as
\begin{equation}
  \begin{array}{lll}
 \displaystyle   \frac{\partial E}{\partial \betaaa_{k,l}^{m}} &=& \displaystyle \sum_{p,q} \frac{\partial E}{\partial \beta_{p,q}^m} . \frac{\partial \beta_{p,q}^{m}}{\partial \betaaa_{k,l}^{m}} \\ 
              \ \ \ \ \textrm{with}   & & \ \ \ \displaystyle  \frac{\partial \beta_{p,q}^{m}}{\partial \betaaa_{k,l}^{m}} =   \displaystyle  \frac{h'(\betaaa_{k,l}^{m})}{\sum_{k',l'} h( \betaaa_{k',l'}^{m})} .  (\delta_{p,q,k,l}-\beta_{p,q}^{m}),                                                                                                                                                                                   \end{array}
                                                                                                                                                                                           \end{equation} 
                                                                                                                                                                                           \noindent and $\delta_{p,q,k,l}=1_{\{(p,q)=(k,l)\}}$. In practice $h(.)=\exp(.)$ and   $\frac{\partial E}{\partial \beta_{p,q}^{m}}$ is obtained from layerwise gradient backpropagation (as already integrated in standard deep learning tools including PyTorch).  Hence,  $\frac{\partial E}{\partial \betaaa_{k,l}^{m}}$ is obtained by multiplying the original gradient $\big[\frac{\partial E}{\partial {\beta}_{p,q}^{m}}\big]_{p,q}$ by the Jacobian  $\big[\frac{\partial \beta_{p,q}^{m}}{\partial \betaaa_{k,l}^{m}}\big]_{p,q,k,l}$ which simply reduces to  $\big[\beta_{k,l}^{m} (\delta_{p,q,k,l}-\beta_{p,q}^{m})\big]_{p,q,k,l}$ when  $h(.)=\exp(.)$.  Similarly, we implement the constraints associated to the appearance stream. \\

                                                                                                                                                                                           \noindent {\bf ResNet update.} As discussed earlier, motion and appearance ResNets are {\it recurrently} (iteratively) applied frame-wise prior to pool the underlying feature maps using multiple aggregation. It is clear that the  number of frames intervening in this aggregation is video-dependent, and thereby the number of terms in these aggregations (and the number of ResNet branches/instances) is also varying. Hence, a straightforward application of the chain rule in the whole architecture -- in order to update  $\frac{\partial E}{\partial \alpha}$ -- becomes possible only when this architecture is unfolded, and this requires fixing the maximum number of frames (denoted as $T$) and sampling temporally all the videos in order to make $T_i$ constant and equal to $T$.  Note that beside requiring all the ResNet instances to share the same parameters (as in Siamese nets), this results into a cumbersome architecture even for reasonable $T$ values.  Furthermore, frame sampling requires interpolation techniques which are highly dependent on quality, duration and temporal resolution of videos and this may result into  spurious motion/appearance details (especially on short videos; even when timely well resolute) which ultimately  leads to a significant drop in action recognition performances. \\ 
                                                                                                                                      
                                                                                                                                      \indent In order to avoid these drawbacks and to fully benefit from the available number (and also temporal resolution) of frames --- without using multiple instances of ``Siamese-like'' ResNets and without resampling --- we consider an alternative gradient estimation. The latter relies on a membership  measure $\mu$ which assigns each frame $f_{i,t}$ to nodes in the temporal pyramid as $\mu_{i,t}^{k,l}={1}_{ \{t \in \N_{k,l}\}}$. Using this membership measure together with the chain rule, the gradient of the loss $E$ w.r.t the parameters of the ResNet $\alpha$ can be updated as 

                                                                                                                                      \begin{equation}
\begin{array}{lll}                                                                                                                    \displaystyle \frac{\partial E}{\partial \alpha_m} &=&  \displaystyle \sum_{k,l} \sum_{i,t} \mu_{i,t}^{k,l} \  \frac{\partial E}{\partial \psi_{k,l}^m}  \  \frac{\partial \psi_{k,l}^m}{\partial \phi_m(f_{i,t})} \  \frac{\partial \phi_m(f_{i,t})}{\partial \alpha_m}.
\end{array}\label{eqgradient} 
\end{equation}
\noindent  Similarly, we evaluate the gradient for the appearance stream. From the above equation, it is clear that when $k=l=1$, all the frames $\{f_{i,t}\}$ contribute in the estimation of the gradient, while for other nodes, only a subsets of frames (belonging to these nodes) are used. Nonetheless, all the frames contribute evenly through all the nodes and hence in gradient estimate, without any sampling. Note also that this formulation implicitly implements {\it weight sharing} as the above gradient can equivalently be written as the sum of gradients, shared through multiple streams of an unfolded architecture, with each stream being dedicated to one frame.  However, the advantage of the above formulation resides again in its computational efficiency and also   its ability to leverage all (possibly variable numbers of) frames in videos while an unfolded architecture requires sampling a fixed number of frames and handling multiple ResNet branches which may clearly lead to intractable training. 

\section{Experiments}
 
In this section, we evaluate the impact of our multiple aggregation design on the performance of action recognition and we compare it against other aggregation strategies as well as the related work using three standard datasets: UCF-101, HMDB-51 and JHMDB-21 \cite{ucf,dataset_HMDB}. UCF-101 ---  used to comprehensively study the different settings of our model --- is the largest and most challenging; it includes 13,320 video shots belonging to 101 categories with variable duration, poor frame resolution, viewpoint and illumination changes, occlusion, cluttered background and eclectic content ranging  from multiple and highly interacting individuals to single and completely passive ones. We also consider HMDB-51 and JHMDB-21 for further comparisons; the latter include 6766 (resp. 928) videos belonging to 51 (resp. 21) action categories. In all these experiments, we process all the videos using ResNet-101 (as a backbone network) in order to extract all the underlying appearance and motion representations framewise. Then, we apply different aggregation schemes prior to assign those videos to classes. We use the same evaluation protocols as the ones suggested in [18, 67, 69] (i.e., train/test splits) and we report the average accuracy over all the categories of actions. \\

\indent \textcolor{black}{We train our complete temporal pyramid-based networks  (in Fig. \ref{fig01}) for respectively 130, 100 and 65 iterations on UCF-101, HMDB-51 and JHMDB-21 using the PyTorch SGD optimizer. For appearance stream, we set the learning rate to 0.001 and reduce it by a factor of 10 every 25, 20, 10 iterations for resp. UCF-101, HMDB-51 and JHMDB-21. For motion stream, we set the learning rate to 0.005 and we reduce it by the same factor after  ``80 and 110'', ``60 and 80'', ``50 and 60'' iterations on the three sets respectively. Experiments on individual streams are run using 4 Titan X Pascal GPUs (with 12 Gb) and last 72h for UCF101, 36h for HMDB-51 and 15h for JHMDB-21 (on the appearance stream) and 96h for UCF101, 48h for HMDB-51 and  24h for JHMDB-21 (on the motion stream) while on the joint stream experiments are run using 4 Tesla P100 GPUs (with 16 Gb) and last 100h, 55h and 30h on the three sets respectively.} 

\begin{table}[h!]
\centering
\resizebox{0.8\linewidth}{!}{
    \centering
    \begin{tabular}{c||c|c}
Deep convolutional networks & UCF-101 & \# parameters (millions) \\
\hline
Pretrained AlexNet \cite{alexnet} & 58.14 & 61M \\
Pretrained VGGNet11 \cite{VGG}& 63.12 & 132M   \\
Pretrained VGGNet19 \cite{VGG} & 63.42   & 143M \\
Pretrained ResNet18 \cite{Resnet16} & 68.32 & 11M\\
Pretrained ResNet50 \cite{Resnet16} & 68.39 & 25M \\
Pretrained ResNet101 \cite{Resnet16} & 68.47 & 44M \\
Pretrained ResNet152 \cite{Resnet16} & 68.58 & 60M \\
    \end{tabular}}
     \caption{\it Action classification performances using the temporal pyramid in \cite{temporalpyramid} combined with different deep network architectures pretrained with ImageNet (these networks were initially designed to extract appearance features).}
    \label{tab:deep_features}
\end{table}
 
\subsection{Convolutional network selection} The choice of the initial pretrained backbone convolutional network -- that operates at the frame-level  --- should consider two factors; its baseline classification performances and the number of training parameters. The latter is particularly crucial  for action recognition  as the size of training data is limited compared to other neighboring tasks (such as image classification) on which these convolutional networks were initially trained.  Hence, in order to select the most appropriate convnet among a collection of existing ones (namely \cite{alexnet,VGG,Resnet16}), we measure the performance of our temporal pyramid based on the design in \cite{temporalpyramid}. The results in Table \ref{tab:deep_features} show that the deeper the network, the better are the performances. However, in our experiments, we consider {\it ResNet-101}, which provides descent action recognition performances  while being relatively less memory and time demanding compared to the other networks and particularly {\it ResNet-152} (see again Table \ref{tab:deep_features}).

\subsection{Settings and Performances}
\indent Firstly, we show a comparison of action recognition performances using different settings. Extensive experiments, reported in Tables.~(\ref{tab:motion01a}) and~(\ref{tab:appearance01a}), show that our hierarchical aggregation design makes it possible to select the best configuration (combination) of level representations in order to improve the performance of classification; indeed, the results show a clear gain as the depth of the hierarchy increases and compared to global average pooling (level~1). This gain results from the match between the temporal granularity of the learned level-wise representations in the hierarchy and the actual granularity of action categories. Note that in all these performances, multi-level node concatenation provides a clear gain compared to averaging, especially on deeper levels of the temporal pyramid, both on motion and appearance streams. The rational is that multi-level node concatenation preserves better the temporal granularity of actions compared to averaging. Hence, in the remainder of these experiments, we keep concatenation when learning ``end-to-end'' joint combination of appearance and motion streams. \\
\begin{table}
\centering 
  \resizebox{0.8\columnwidth}{!}{
    \begin{tabular}{c||cc|cc}
        \bf Motion stream    & \multicolumn{2}{c}{Shallow design} & \multicolumn{2}{c}{Deep design} \\
      UCF-101 & concatenation & averaging & concatenation & averaging  \\
      \hline
      \hline
                  TP (level 1) &  78.40 &78.40  &\bf 78.66 & 78.66 \\
                  TP (level 2)   & 79.53 & 79.54 &\bf 79.86  & 79.76  \\
                  TP (level 3) &79.70  &79.71 & \bf 79.93 & 79.83 \\
                  TP (level 4) & 79.76 &79.77 & \bf 81.14 &80.66 \\
                  TP (level 5)   &80.23 & 80.24  & \bf 81.43 &80.84    \\
                  TP (level 6) &79.96 & 79.98  & \bf 81.69 &  80.12    
    \end{tabular}}
  \caption{\it \textcolor{black}{This table shows level-wise performances using the motion stream both for shallow and deep models. These performances are reported both for ``averaging'' and ``concatenation''. In these initial experiments -- in order to compare the performances of shallow and deep designs under comparable conditions -- we fine-tune only the last fully connected layer of {\it ResNet-101} along with the parameters of the temporal pyramid (TP).}}
    \label{tab:motion01a}
  \end{table}
  \begin{table}[h!]
\centering
    \resizebox{0.8\columnwidth}{!}{
    \begin{tabular}{c||cc|cc}
        \bf Appear stream    & \multicolumn{2}{c}{Shallow design} & \multicolumn{2}{c}{Deep design} \\
      UCF-101 & concatenation & averaging & concatenation & averaging  \\
      \hline
      \hline
                  TP (level 1) & 80.28  & 80.28 &\bf 80.31  & 80.31  \\
                  TP (level 2) &81.77 & 81.78 & 82.16  &  \bf 82.21 \\
                  TP (level 3) &82.17 &82.17 & 82.74  & \bf 82.89 \\
                  TP (level 4)  & 82.51 &82.50 & \bf 83.52 &83.38  \\
                  TP (level 5)&82.50 &  82.51 & \bf 83.63   & 80.83 \\
                  TP (level 6) &81.96 & 81.96 &\bf 83.92  &  80.83                                                         
    \end{tabular}
    }
    \caption{\it \textcolor{black}{This table shows level-wise performances using the appearance stream both for shallow and deep models. These performances are reported both for ``averaging'' and ``concatenation''. In these initial experiments -- in order to compare the performances of shallow and deep designs under comparable conditions -- we fine-tune only the last fully connected layer of {\it ResNet-101} along with the parameters of the temporal pyramid.}}
    \label{tab:appearance01a}
    \end{table}
\begin{table}[h!]
      \centering
\resizebox{0.95\columnwidth}{!}{
    \begin{tabular}{c||ccc||ccc|cc}
        \bf Fusion   & \multicolumn{3}{c}{Shallow design (concat) } & \multicolumn{3}{c}{Deep design (concat)} & \multicolumn{2}{c}{Stream importance}  \\
      UCF-101 & Motion & Appear & Joint & Motion &  Appear & Joint  & $\w_m$ & $\w_a$ \\
      \hline
      \hline
                  TP (level 1)  &78.40   &80.28 & 88.91 & 78.74  &80.69  & 89.69 & 0.46 & \bf 0.54 \\ 
                  TP (level 2)  &79.53   &81.77 & 89.10 &79.97  &82.78 & 90.00& 0.49 & \bf 0.51 \\ 
                  TP (level 3)  &79.70  & 82.17 & 89.34 &80.69  & 83.12 &  90.26 & \bf 0.52 & 0.48 \\ 
                  TP (level 4)  &79.76   &82.51 & 89.37 &81.74  &83.78 &90.92 & \bf 0.52 &0.48 \\ 
                  TP (level 5)  &80.23   &82.50 & 84.49 &82.86  &84.10 & 91.45 & \bf 0.56 & 0.44\\ 
                  TP (level 6)  &79.96  &81.96 & 89.26 & 83.41  &84.92  & \bf 92.37 & \bf 0.60 & 0.40 

    \end{tabular}}
    \caption{\it This table shows level-wise  performances of joint (2-stream) fusion for both shallow and deep methods. These results are shown only for ``concatenation'' as the underlying baseline performances reported in tables.~\ref{tab:motion01a} and~\ref{tab:appearance01a} are better than ``averaging''. In contrast to tables (\ref{tab:motion01a}) and  (\ref{tab:appearance01a}), all the parameters of the whole network (including ResNet) are allowed to vary.}
    \label{tab:late_fusionaa}
    \end{table}
    
\begin{figure}[h!]
  \centering
\resizebox{0.4\columnwidth}{!}{
  \includegraphics[width=0.28\columnwidth]{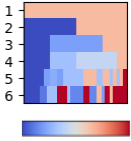}
  \includegraphics[width=0.29\columnwidth]{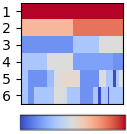}}
\begin{center} (a) Single temporal pyramid    \end{center}
\resizebox{0.7\columnwidth}{!}{
\includegraphics[width=0.99\columnwidth]{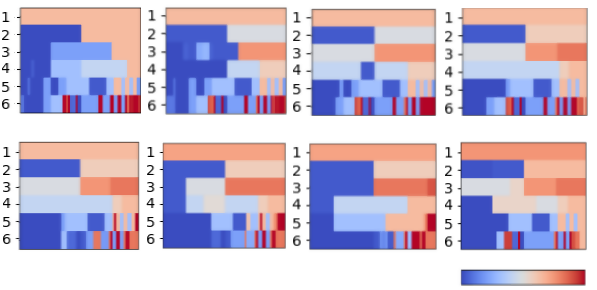}}
\begin{center} (b) Multiple temporal pyramids (motion stream)   \end{center} 
\resizebox{0.7\columnwidth}{!}{
    \includegraphics[width=0.99\columnwidth]{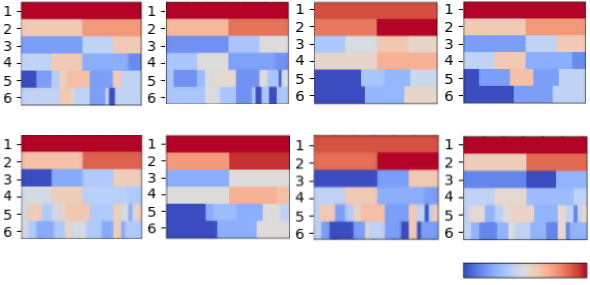}}
 \begin{center} (c) Multiple temporal pyramids (appearance stream)   \end{center}
    \caption{\it  (a) Weight distribution of motion and appearance streams obtained when learning the parameters of a single temporal pyramid (corresponding to the first row of Table \ref{tab:my_label}). (b-c) Weight distribution of multiple temporal pyramids of motion and appearance streams (corresponding to the fourth row in the same table). Warmer colors correspond to higher weights while cooler colors to lower ones.}
   \label{fig:heatmaps_ConTem Pyramid}
\end{figure}

    \indent Secondly, we compare the performance of the two settings (shallow and deep) of our multiple aggregation design using both motion and appearance streams taken individually and combined; as already discussed, the parameters  $\w_a$, $\w_m$ of this fusion are optimized as a part of the end-to-end learning process. \textcolor{black}{Results reported in Table \ref{tab:late_fusionaa} show the complementary aspects of the two streams in all the settings} as their fusion brings a clear gain in performance. Moreover, we observe that the contribution of the motion stream is strictly increasing (and  {\it a contrario} strictly decreasing for appearance stream) as the level of the temporal pyramid increases \textcolor{black}{(see the distribution of $\w$ in Table \ref{tab:late_fusionaa})}. This clearly corroborates the highest impact of motion (compared to appearance) when modeling the temporal granularity of action categories (see later Fig.~\ref{fig:heatmaps_ConTem Pyramid}). We also observe a higher positive impact on performances as the depth of our temporal pyramids increases; again, these results are obtained using ``concatenation'' instead of ``averaging'', as the former  already globally overtakes the latter on motion and appearance streams when taken individually (see again Tables~\ref{tab:motion01a} and~\ref{tab:appearance01a}). \\
    
\begin{table}[h!]
\centering
  \resizebox{0.8\columnwidth}{!}{
    \begin{tabular}{c|ccc}
         \# of temporal   & \multicolumn{3}{|c}{Accuracy (concatenation)}\\
         pyramids per stream & Appearance stream & Motion stream & Joint stream \\
      \hline
      \hline
         1 & 83.92 & 81.69 & 90.78 \\
         2 & 83.95 & 81.73  &90.79  \\
         4 &  \bf 83.97& 81.79 & 90.84 \\
         8 &  83.92& \bf 81.86 & \bf 90.89\\
         16 &  83.89& 81.83 & 90.85 
    \end{tabular}}
    \caption{\it This table shows the evolution of the performances w.r.t different \# of temporal pyramids per stream. In order to combine the outputs of these multiple pyramids (when using concatenation), we add a succession of FC+ReLU+BatchNorm to reduce the dimensionality {\bf from} ``63 (number of nodes in TP of 6 levels) $\times$ 128 (node dimension) $\times$ \# TPs'' {\bf to} ``128''. All these results correspond to temporal pyramids of 6 levels.}
    \label{tab:my_label}
  \end{table}
    \indent We further investigate the potential of our method using multiple instances of temporal pyramids both for motion and appearance streams as well as their joint fusion. The rational -- from this setting --  resides in the heterogeneity of action categories and their dynamics which may affect the accuracy; indeed, the apex of some actions appears early in video clips while for others later or spread through all the video duration. Hence, instead of learning a single monolithic temporal pyramid per stream, we stack multiple instances of temporal pyramids with different weights $\beta$, each one dedicated to a subclass of actions whose dynamics (not category) are similar\footnote{These subclasses of actions are not explicitly defined in a supervised manner but implicitly by allowing enough flexibility in the multiple instances of temporal pyramids in order to capture different (unknown) subclasses of action dynamics.}. We learn the parameters of these pyramids ``end-to-end'' as discussed earlier for single pyramids.  Table~\ref{tab:my_label} shows the performances w.r.t the number of pyramids. In spite of an increase of the number of training parameters in these multiple pyramids (without any increase of training data), we observe an improvement; we believe that adding extra training data will bring a further and clearer gain in performances. 
\begin{table}[h!]
\centering
  \resizebox{0.8\linewidth}{!}{
    \begin{tabular}{c|cc|cc||ccc}
    Sampling & \multicolumn{2}{|c|}{\# frames (train)} & \multicolumn{2}{|c||}{\# frames (test)}& \multicolumn{3}{|c}{Accuracy} \\
     strategies                    & RGB & OF  &RGB & OF & Appearance & Motion & Fusion\\
     \hline    \#1 & 25 & 25  & 25 &25 &84.23&81.27&91.65 \\
     \#2 & 25 & 25  & 25 &250 &84.23&81.27&91.64 \\
      \#3 & 25 & 50  & 25 &50 &84.23&81.86&91.69  \\
      \#4 & 25 & 50  & 25 &250 &84.23&81.89&91.78 \\
      \#5 & 64 & 64  & 250 &250 &84.62&82.05&91.89 \\
      \#6 & 64 & 64  & all &all &84.81&82.77&92.09 \\
     \#7 & 64 & all  & all &all &84.81&83.41&92.29 \\
      \#8 & all & all &  all &all &84.92&83.41 & 92.37 
    \end{tabular}}
    \caption{\it This table shows the evolution of the performance w.r.t to different sampling strategies (i.e., number of frames in training and test videos). RGB and OF stand for the number of input RGB frames and the  number of optical flow frames used in the appearance and the motion streams respectively. These performances are obtained using a temporal pyramid of six levels.}
    \label{tab:sampling_parametera}
\end{table}

\subsection{Sampling, surrogate gradient and efficiency}
 
Table.~\ref{tab:sampling_parametera}  shows the impact of our method  -- with and without frame sampling --  on the performance of action recognition. These results are obtained using a single pyramid.  From these results, it is easy to see that performances get better as the number of sampled frames increases reaching asymptotically the best performances when all the frames are used. This behavior is similar both on motion and appearance streams. \textcolor{black}{However, we notice that motion stream which is based on optical flow data is more sensitive to sampling than appearance stream so the accuracy of the former is clearly proportional to the number of frames. Put differently, motion stream builds a better representation and hence becomes more important for the overall action classification when it is fed with more optical flow data as shown again in Table~\ref{tab:sampling_parametera} (settings \#6 and \#7).} However, taking all the frames during backpropagation, comes at the expense of a substantial increase of computation; when considering all the 2.5 millions frames of our videos on UCF-101, training costs 72h (resp. 96h) for appearance (resp. motion) stream using 4 Titan X GPUs (with 12 Gb) and 100h on the joint stream using 4 Tesla P100 GPUs (with 16 Gb). This high cost results from the large number of visited frames when (re)estimating the gradient, in Eq.~\ref{eqgradient} w.r.t the parameters of the ResNet, through the epochs of backpropagation. In order to make the evaluation of Eq.~\ref{eqgradient} (and hence training) more tractable (with a controlled loss  in classification performances), we consider a surrogate gradient defined as

\begin{equation} 
\begin{array}{lll}                                                                                                                    \displaystyle \frac{\partial E}{\partial \alpha_m} &=&  \displaystyle \sum_{k,l,i} \ \  \sum_{t \in {\cal P}_r^i} \mu_{i,t}^{k,l} \  \frac{\partial E}{\partial \psi_{k,l}^m}  \  \frac{\partial \psi_{k,l}^m}{\partial \phi_m(f_{i,t})} \  \frac{\partial \phi_m(f_{i,t})}{\partial \alpha_m},
\end{array}\label{eqgradient2} 
\end{equation}

\noindent here ${\cal P}_r^i$ stands for a subset of selected frame time-stamps, in a given video $\V_i$, that contribute to gradient estimation at the $r^{\textrm{th}}$ epoch. We consider a periodic selection mechanism which guarantees that all the frames are evenly used through epochs; in practice,  ${\cal P}_r^i= \{ t \in [0,T_i], \ t \equiv r \pmod  K\}$ with $1\slash K$ being the fraction of frames used per epoch.  With this mechanism, gradient evaluation still relies on the entire set of frames in the training set, but their use is distributed through epochs and this makes the evaluation and training process far more efficient while maintaining close performances (see Table.~\ref{tab:surrogate_back_propa}). For instance, when $K=24$, training is  $24\times$ faster compared to the most accurate setting (strategy \#8 in Table~\ref{tab:sampling_parametera}) as only 8 frames are used (on average ``per epoch-per video'')  in Eq.~\ref{eqgradient2} instead of 185; furthermore, since all the frames contribute equally through all the epochs, the loss in  accuracy is contained. These performances are obtained on individual and joint streams using   the same aforementioned hardware resources. 

\begin{table}[h!]
  \centering
 \resizebox{0.99\linewidth}{!}{
    \begin{tabular}{c||c|ccc}
    Speed up    & & \multicolumn{3}{c}{Accuracy}  \\
    factor (K) & Avg. \#  frames per "epoch and training video" &  Appearance & Motion & Joint \\ 
      \hline
      \hline 
      $1 \times$ & 185&  84.92 &  83.41& 92.37 \\
     $4 \times$ & 92 &  84.27& 82.59 &91.74  \\
    $8 \times$ & 46 & 84.10 &82.07 &91.39 \\
    $16 \times$ & 23 & 83.96 & 81.23&90.70 \\
     $24\times$ & 8 &  83.89&80.95& 90.35  \\               
    \end{tabular}}
    \caption{\it This table shows the performance of ``surrogate back-propagation'' with different acceleration factors.  Note that motion stream performances are more sensitive to this acceleration compared to appearance stream.} 
    \label{tab:surrogate_back_propa}
  \end{table}
  \begin{table*}[h!]
    \centering
    \resizebox{0.99\linewidth}{!}{
    \begin{tabular}{c||c|c|c|c|c|c|c}
Methods & \rotatebox{90}{UCF-101}& \rotatebox{90}{HMDB-51} & \rotatebox{90}{JHMDB-21} & \rotatebox{90}{Batch size}& \rotatebox{90}{\# frames (RGB,OF)} & \rotatebox{90}{ImageNet pretrain} &\rotatebox{90}{Kinetics pretrain} \\
    \hline
2D colorized heatmaps~\cite{pose} &64.38&54.90&60.5 &32 &(all,all)&\xmark&\xmark \\
\hline
2D motion + GAP  \cite{pretrained_resnet101_ucf}&79.4 &59.13 &61.39&32&(none,64)&\cmark&\xmark \\ 
2D appearance + GAP \cite{pretrained_resnet101_ucf}&82.1 &60.24&62.71&32&(3,none) &\cmark&\xmark \\ 
2D 2-streams +  GAP \cite{pretrained_resnet101_ucf}&88.5 &63.31 &64.11&32&(3,64)&\xmark&\xmark \\  
\hline
3D motion \cite{kin3d} &96.41&80.39&\xmark& 15&(none,64)&\cmark&\cmark  \\  
3D appearance \cite{kin3d} &95.60&76.47& \xmark & 15&(64,none)&\cmark&\cmark \\ 
3D two-streams \cite{kin3d} &97.94 &80.65 &\xmark & 15&(64,64)&\cmark&\cmark \\ 
        \hline
TP-A of~\cite{temporalpyramid} (on ResNet152~\cite{Resnet16})&68.58 &58.63 &62.16 &\xmark &(all,none)&\cmark&\xmark \\
Spect-A (on ResNet152+ResNet18~\cite{Resnet16})& 64.41& 54.85&60.61 &32&(all,all)&\cmark&\xmark \\
Spect-A (on ResNet101+ResNet18 \cite{Resnet16})& 78.40& 57.76&61.26&32&(all,all)&\cmark&\xmark \\  
Spect-M (on ResNet101+Resnet18 \cite{Resnet16})& 76.46& 55.38&60.66&32&(all,all)&\cmark&\xmark \\  
Spect-2S (on ResNet101+Resnet18 \cite{Resnet16}) & 80.10& 58.28&62.14&32&(all,all)&\cmark&\xmark \\ 
      \hline
      \hline
      Our "2D motion + TP"  &83.41 &61.04 &62.97&1&(all,all)&\cmark&\xmark \\ 
Our "2D appearance + TP"&84.92 &62.23 &63.51&1&(all,all)&\cmark&\xmark \\  
Our "2D two-streams + TP"  &92.37 &65.14 &66.96&1&(all,all)&\cmark&\xmark \\ 
        \hline
2D col-heatM\cite{pose} + our "2D motion + TP" & 80.41 &65.21&69.93 &\xmark&\xmark&\xmark&\xmark \\ 
3D motion\cite{kin3d} + our "2D motion + TP" &96.61 &80.54 &\xmark&\xmark&\xmark&\xmark&\xmark \\  
3D appear\cite{kin3d} + our "2D appear + TP" &96.05 &76.56&\xmark&\xmark&\xmark&\xmark&\xmark  
    \end{tabular}}
    \caption{\it This table shows a comparison of our temporal pyramid (TP) w.r.t different related works; in this table, ``col-heatM'' stands for colorized heatmaps, ``Spect'' for spectrograms, ``A'' for appearance, ``M'' for motion, ``2S'' for two-streams, ``GAP'' for global averge pooling and ``OF'' for optical flow. In our experiments, (i) {\it ResNet-152} is pretrained on ImageNet, (ii) {\it ResNet-101} is pretrained on ImageNet and fine-tuned on UCF-101 (for both appearance and motion) and (iii) ResNet18 is pretrained on ImageNet and fine-tuned on  UCF-101 (again for appearance and motion). In these results, the symbol "\xmark" stands for "a method does not apply or was not applied (results not available)" in the underlying works.}
    \label{comparsion_state_of_art_methoda}
\end{table*}
\subsection{Comparison against related work}
\indent Finally, we compare the performance and the complementary aspects of our method against related state of the art action recognition methods \cite{pose,pretrained_resnet101_ucf,temporalpyramid,kin3d,Resnet16} on  UCF-101, HMDB-51 and JHMDB-21. The closely related method in~\cite{temporalpyramid} is based on deep framewise representations which are aggregated and classified using a hierarchy of multiple temporal granularities. However, the method in~\cite{temporalpyramid} differs from the one proposed in this paper in different aspects: first, framewise representations are extracted using ResNet-152 pretrained only on ImageNet and not fine-tuned on UCF-101. Besides, the method in~\cite{temporalpyramid} is based only on appearance stream and more importantly, the design principle of our proposed method is deep and consists in weighting the contribution of each level in the temporal pyramid as a part of an ``end-to-end'' learning process while in ~\cite{temporalpyramid} this weighting scheme is relatively shallow and excludes the ResNet from training. All these differences explain the significant under-performances of \cite{temporalpyramid} compared to our method as shown in Table.~\ref{comparsion_state_of_art_methoda}.  \\
\indent Extra comparisons in Table.~\ref{comparsion_state_of_art_methoda} also include global averaging techniques as well as spectrogram-like representations. The former produces a global representation that averages all the frame representations while the latter keeps all the frame representations and concatenate them prior to their classifications.  Note that these two settings are related to the two extreme cases of our hierarchy, i.e., the root and the leaves.  In particular, the spectrogram of a video $\V$ with $T$ frames is obtained when the number of leaf nodes, in the hierarchy, is exactly equal to $T$. Global averaging techniques (shown in Table.~\ref{comparsion_state_of_art_methoda}) include \cite{pose}; the latter is based on colorized heatmaps and corresponds to timely-stamped and averaged framewise probability distributions of human keypoints. These colorized heatmaps are fed to a 2D CNN for classification; note that colorized heatmaps provide video-level representations which capture globally the dynamics of video actions without any scheme to emphasize the most important temporal granularities of these actions and this results into low accuracy as again displayed  in Table.~\ref{comparsion_state_of_art_methoda}.\\

\indent The last category of methods (shown in Table.~\ref{comparsion_state_of_art_methoda}) include convolutional networks based on 2D and 3D spatio-temporal filters~\cite{pretrained_resnet101_ucf,kin3d}. These methods are based either on one or two streams; one for motion and another one for appearance  followed by a global average pooling. Both methods are similar to ours; they combine motion and appearance streams and their design is end-to-end but clearly differ in their pooling mechanisms and the way frames are exploited. Indeed, these related techniques rely on sampling strategies that vectorize video sequences into fixed length inputs while our method keeps all the frames in order to build temporal pyramids. Another major difference w.r.t our method resides in the huge set used in order to train these related architectures. Nevertheless, while these streams are highly effective their combination with our hierarchical aggregation, through a late fusion\footnote{Late fusion is applied (instead of early one) as our video inputs are different from those of 2D colorized heatmaps and convolutional 3D filters which are spatio-temporal while ours are only spatial. We also exclude, from fusion, 2D methods+GAP as they correspond to a particular setting of our method (namely temporal pyramid of level 1).}, brings a noticeable gain in performances. We also observe the same behavior on all the combinations of our two stream model with other baselines and other related methods (including two stream 3D CNNs \cite{kin3d} and spectrograms \cite{temporalpyramid}); indeed, from the results shown in Table.~\ref{comparsion_state_of_art_methoda}, our hierarchical method brings a clear gain w.r.t most of these methods. Note that some of these models rely on extra datasets (including Kinetics) in order to pretrain their CNNs while our method is trained only on the original datasets. 

\section{Conclusion}

We introduce in this paper a temporal pyramid approach for video action recognition. The strength of the proposed method resides in its ability to learn hierarchical pooling operations that capture different levels of temporal granularity in action recognition. This is translated into learning the distribution of weights in the temporal pyramid, that capture these granularities, by solving constrained minimization problems. Two settings are considered: shallow and deep.  The former relies on solving a constrained quadratic programming problem while the latter on optimizing the parameters of a deep network including a temporal pyramid module both on motion and appearance streams as well as their combination. We also consider variants of the deep learning framework that designs multiple instances of temporal pyramids each one dedicated to a particular subcategory of action granularities and also a procedure that allows us to efficiently train the network at the detriment of a slight decrease of its classification accuracy. The advantages of these contributions are established, against different baselines as well as the related work, through extensive experiments on challenging action recognition benchmarks including UCF-101, HMDB-51 and JHMDB-21 datasets.\\
As a future work, we are currently investigating the issue of learning other aggregation schemes, besides hierarchical averaging and concatenation, and also the extension of this method to other benchmarks and other visual recognition tasks. 
 
{
   
}


\begin{thebibliography}{9}
   \bibitem{kin3d}
 J. Carreira, A. Zisserman. Quo Vadis, Action Recognition? A New Model and the Kinetics Dataset. In IEEE Conference on Computer Vision and Pattern Recognition (CVPR), 2017
   \bibitem{scene03}
M. Pantic, A. Pentland, A. Nijholt, T.S. Huang. Human Computing and Machine Understanding of Human Behavior: A Survey. In Human Computing and Machine Understanding of Human Behavior, 2007




\bibitem{sahbiicpr18}
H. Sahbi  and N. Boujemaa. "From coarse to fine skin and face detection." Proceedings of the eighth ACM international conference on Multimedia. 2000.

   \bibitem{temporalpyramid_detec}
 H. Pirsiavash, D. Ramanan. Detecting Activities of Daily Living in First-person Camera Views. In IEEE Conference on Computer Vision and Pattern Recognition (CVPR), 2012


 \bibitem{mkl_action}
 L. Chen, L. Duan, D. Xu. Event Recognition in Videos by Learning From Heterogeneous Web Sources. In IEEE International Conference on Computer Vision and Pattern Recognition(CVPR), 2013
 
 \bibitem{mklimage2017}
M. Jiu, H. Sahbi. Nonlinear deep kernel learning for image annotation. IEEE Transactions on Image Processing, volume 26, number 4, 1820-1832, 2017.

\bibitem{sahbiicassp16b}
M. Jiu, H. Sahbi. Laplacian deep kernel learning for image annotation. IEEE International Conference on Acoustics, Speech and Signal Processing, 2016. 

  \bibitem{temporal_pyramid}
 D. Xu, S-F. Chang. Visual Event Recognition in News Video using Kernel Methods with Multi-Level Temporal Alignment. In IEEE International Conference on Computer Vision and Pattern Recognition(CVPR), 2007

\bibitem{superived_dic_action}
H. Wang, C. Yuan, W. Hu, C. Sun. Supervised class-specific dictionary learning for sparse modeling in action recognition. Pattern Recognition, Volume 45, Issue 11, Pages 3902-3911, 2012
 

\bibitem{multi_svm}
 C. Schuldt, I. Laptev, B. Caputo. Recognizing human actions: a local SVM approach. In IEEE International Conference on Pattern Recognition (ICPR), 2004
  \bibitem{sahbiigarss12b}
N. Bourdis, D. Marraud and H. Sahbi. "Spatio-temporal interaction for aerial video change detection." 2012 IEEE International Geoscience and Remote Sensing Symposium. IEEE, 2012.

  \bibitem{spresnet16}
 C. Feichtenhofer, A. Pinz, R-P. Wildes. Spatiotemporal Residual Networks for Video Action Recognition. In Neural Information Processing Systems (NeurIPS), 2016
 
 \bibitem{spresnetmulti17}
 C. Feichtenhofer, A. Pinz, R-P. Wildes. Spatiotemporal Multiplier Networks for Video Action Recognition. In IEEE International Conference on Computer Vision and Pattern Recognition(CVPR), 2017 

   \bibitem{pose}
     M. Liu, Y.  Junsong. "Recognizing human actions as the evolution of pose estimation maps." Proceedings of the IEEE Conference on Computer Vision and Pattern Recognition. 2018.

\bibitem{sahbijmlr06}
H. Sahbi, D. Geman. A hierarchy of support vector machines for pattern detection. Journal of Machine Learning Research 7.Oct (2006): 2087-2123.
 
  \bibitem{segment_net}
  L. Wang, Y. Xiong, Z. Wang, Y. Qiao, D. Lin, X. Tang, L. Van Gool. Temporal Segment Networks: Towards Good Practices for Deep Action Recognition. In European Conference on Computer Vision (ECCV), 2016
    \bibitem{MKL}  M. Gönen, E. Alpaydın. Multiple Kernel Learning Algorithms. In Journal of Machine Learning Research (JMLR) : 2211-2268, 2011
 \bibitem{sahbispie2004}
N. Boujemaa, J. Fauqueur, M. Ferecatu, F. Fleuret, V.  Gouet, B. L. Saux, and H.  Sahbi. "Ikona: Interactive generic and specific image retrieval." In Proceedings of the International workshop on Multimedia Content-Based Indexing and Retrieval (MMCBIR?2001), pp. 25-29. 2001.

   \bibitem{ucf}
K. Soomro, A-R. Zamir and M. Shah. UCF101: A Dataset of 101 Human Action Classes From Videos in The Wild, CRCV-TR-12-01, November, 2012.

  \bibitem{MKL_alignement}
  C.cortes, M. Mohri, A. Rostamizadeh. Algorithms for learning Kernels based on Centered Alignement. In Journal of Machine Learning Research (JMLR) : 795-828, 2012

  \bibitem{art_imagenet}
  B. Zoph, V. Vasudevan, J. Shlens, Q-V. Le. Learning Transferable Architectures for Scalable Image Recognition.  In IEEE International Conference on Computer Vision and Pattern Recognition(CVPR), 2018
\bibitem{icip2001}
H. Sahbi and N. Boujemaa. "Robust matching by dynamic space warping for accurate face recognition." Proceedings 2001 International Conference on Image Processing (Cat. No. 01CH37205). Vol. 1. IEEE, 2001.

   \bibitem{imagenet}
J. Deng, W. Dong, R. Socher, L.-J. Li, K. Li and L. Fei-Fei, ImageNet: A Large-Scale Hierarchical Image Database. In IEEE International Conference on Computer Vision and Pattern Recognition(CVPR), 2009. 
 \bibitem{STP_CV17}
Y. Wang, M. Long, J. Wang, Philip S. Yu. Spatio temporal Pyramid Network  for  Video  Action  Recognition. In IEEE International Conference on Computer Vision and Pattern Recognition(CVPR), 2017

\bibitem{sahbiclef08}
S. Tollari, P. Mulhem, M. Ferecatu, H. Glotin, M. Detyniecki, P. Gallinari, H. Sahbi,  Z-Q. Zhao. A comparative study of diversity methods for hybrid text and image retrieval approaches. In Workshop of the Cross-Language Evaluation Forum for European Languages, pp. 585-592. Springer, Berlin, Heidelberg, 2008.

 \bibitem{tp_segment}
 J. Zhu, W. Zou, Z. Zhu. End-to-end Video level Representation Learning for Action Recognition. In International Conference on Learning Representation (ICLR), 2018
\bibitem{stp_cnn} 
 Z. Zheng, G. An, D. Wu, Q. Ruan. Spatial-temporal pyramid based Convolutional Neural Network for action recognition. Neurocomputing, Volume 358, 17 September 2019, Pages 446-455
   \bibitem{sahbiicip09}
M. Ferecatu, H. Sahbi. Multi-view object matching and tracking using canonical correlation analysis. 16th IEEE International Conference on Image Processing (ICIP), 2109-2112, 2009. 


 \bibitem{tp_concat}
Zhang D., Dai X., Wang YF. Dynamic Temporal Pyramid Network: A Closer Look at Multi-scale Modeling for Activity Detection. In Asian Conference on Computer Vision (ACCV), 2018
 \bibitem{sahbijstars17}
Q. Oliveau, H. Sahbi. Learning attribute representations for remote sensing ship category classification.  IEEE Journal of Selected Topics in Applied Earth Observations and Remote Sensing, 2017. 

\bibitem{tp_scale}
K. Yang, R. Li, P. Qiao, Q. Wang, D. Li, Y. Dou. Temporal Pyramid Relation Network For Video-based Gesture Recognition. In IEEE International Conference on Image Processing (ICIP), 2018

\bibitem{video_capt01}
B. Wang, L. Ma, W. Zhang, W. Liu. Reconstruction
network for video captioning. In IEEE International Conference on Computer Vision and Pattern Recognition(CVPR), 2018

\bibitem{temporalpyramid}
A. Mazari, H. Sahbi. Deep Temporal Pyramid Design for Action Recognition. In IEEE International Conference on Acoustics, Speech and Signal Processing (ICASSP), 2019
\bibitem{Resnet16}
H. Kaiming, Z. Xiangyu, R. Shaoqing; S. Jian. Deep Residual Learning for Image Recognition. In IEEE International Conference on Computer Vision and Pattern Recognition(CVPR), 2016
    \bibitem{of}
  B. K.P.Horn, B. G.Schunck. Determining optical flow. Artificial Intelligence, Volume 17, Issues 1–3, Pages 185-203, 1981 
  \bibitem{hog}
W. Lu and James J. Little. Simultaneous tracking and action recognition using the pca-hog descriptor. In European conference on Computer vision (ECCV), 2006

\bibitem{stips}
I. Laptev. On Space-Time Interest Points. In International Journal of Computer Vision (IJCV), Volume 64, Issue 2–3, pp 107–123, 2005
\bibitem{scene01}
T. Bagautdinov, A. Alahi, F. Fleuret, P. Fua, S. Savarese. Social Scene Understanding: End-To-End Multi-Person Action Localization and Collective Activity Recognition. In IEEE International Conference on Computer Vision and Pattern Recognition(CVPR), 2017
\bibitem{scene02}
J. Shao, K. Kang, C. Change Loy, X. Wang. Deeply Learned Attributes for Crowded Scene Understanding. In IEEE International Conference on Computer Vision and Pattern Recognition(CVPR), 2015

\bibitem{sahbiigarss12}
N. Bourdis, D. Marraud and H.  Sahbi. "Camera pose estimation using visual servoing for aerial video change detection." 2012 IEEE International Geoscience and Remote Sensing Symposium. IEEE, 2012.



\bibitem{video_surv01}
A. Ben Mabrouk, E. Zagrouba. Abnormal behavior recognition for intelligent video surveillance systems: A review. In Expert Systems with Applications Volume 91, Pages 480-491, 2018


\bibitem{video_surv02}
Y. Han, P. Zhanga, T. Zhuob, W. Huang, Y. Zhanga. Going deeper with two-stream ConvNets for action recognition in video surveillance. In Pattern Recognition Letters Volume 107, Pages 83-90, 2018 

\bibitem{sahbiaccv2010}
H. Sahbi, J-Y. Audibert, and R. Keriven. "Context-dependent kernels for object classification." IEEE transactions on pattern analysis and machine intelligence 33.4 (2010): 699-708.
\bibitem{sahbiijmir15}
H Sahbi. Imageclef annotation with explicit context-aware kernel maps. International Journal of Multimedia Information Retrieval 4 (2), 113-128

\bibitem{video_capt02}
J. Wang, W. Jiang, L. Ma, W. Liu, Y. Xu. Bidirectional attentive fusion with context gating for dense video captioning. In IEEE International Conference on Computer Vision and Pattern Recognition(CVPR), 2018
\bibitem{video_capt03}
N. Aafaq, N. Akhtar, W. Liu, S. Zulqarnain Gilani, A. Mian. Spatio-Temporal Dynamics and Semantic Attribute Enriched Visual Encoding for Video Captioning. In the IEEE Conference on Computer Vision and Pattern Recognition (CVPR), 2019
\bibitem{sahbiclef13}
H. Sahbi, L.  Ballan, G. Serra, and A. Del Bimbo. "Context-dependent logo matching and recognition." IEEE Transactions on Image Processing 22, no. 3 (2012): 1018-1031.

\bibitem{video_capt04}
Minlong Lu, Ze-Nian Li, Yueming Wang, Gang Pan. Deep Attention Network for Egocentric Action Recognition. In IEEE Transactions on Image Processing, Volume 28, Issue 8, 2019
\bibitem{video_capt05}
T. Mahmud, M. Billah, M. Hasan, Am. K. Roy-Chowdhury. Captioning Near-Future Activity Sequences. In arXiv:1908.00943, 2019
\bibitem{vid_retri01}
I. Laptev, P. Perez. Retrieving actions in movies. In International Conference on Computer Vision (ICCV), 2007
\bibitem{vid_retri02}
L. Ballan, M. Bertini, A. Del Bimbo, L. Seidenari, G. Serra. Event detection and recognition for semantic annotation of video. In Multimedia Tools and Applications,  Volume 51, Issue 1, pp 279–302, 2011
\bibitem{sahbisc7}
H. Sahbi.  "A particular Gaussian mixture model for clustering and its application to image retrieval." Soft Computing 12.7 (2008): 667-676.

\bibitem{vid_retri04}
A. Jaimes, K. Omura, T. Nagamine, K. Hirata. Memory Cues for Meeting Video Retrieval. In CARPE Proceedings of the 1st ACM workshop on Continuous archival and retrieval of personal experiences, Pages 74-85, 2004 
\bibitem{sahbicassp11}
  X. Li, H. Sahbi. Superpixel-based object class segmentation using conditional random fields. IEEE International Conference on Acoustics, Speech and Signal Processing (ICASSP). 2011



\bibitem{vid_retri05}
O. Duchenne, I. Laptev, J. Sivic, F. Bach, J. Ponce. Automatic Annotation of Human Actions in Video. In International Conference on Computational Vision (ICCV), 2009
\bibitem{fuzzy05}
H. Sahbi and N. Boujemaa. "Validity of fuzzy clustering using entropy regularization." The 14th IEEE International Conference on Fuzzy Systems, 2005. FUZZ'05.. IEEE, 2005.

\bibitem{vid_robotics01}
H. Meng, N. Pears, C. Bailey. A Human Action Recognition System for Embedded Computer Vision Application. In IEEE Conference on Computer Vision and Pattern Recognition (CVPR), 2007
\bibitem{lingsahbi2013}
 L. Wang, H. Sahbi. Directed Acyclic Graph Kernels for Action Recognition. Proceedings of the IEEE International Conference on Computer Vision. 2013.

\bibitem{vid_robotics02}
T. Theodoridis, A. Agapitos, H. Hu, S.M. Lucas. Ubiquitous robotics in physical human action recognition: A comparison between dynamic ANNs and GP. In IEEE International Conference on Robotics and Automation, 2008
 

\bibitem{vid_robotics03}
Y. Demiris. Prediction of intent in robotics and multi-agent systems. Cogn Process (2007) 8: 151. https://doi.org/10.1007/s10339-007-0168-9
\bibitem{vid_robotics04}
M. Nan, A. Stefania Ghiță, A. Gavril, M. Trascau, A. Sorici, B. Cramariuc, A. Magda Florea. Human Action Recognition for Social Robots. In nternational Conference on Control Systems and Computer Science, 2019
\bibitem{vid_robotics05}
E. Coupeté, F. Moutarde, S. Manitsaris. Multi-users online recognition of technical gestures for natural human–robot collaboration in manufacturing. Robot (2019) 43: 1309. https://doi.org/10.1007/s10514-018-9704-y
\bibitem{lingsahbieccv2014}
 L. Wang, H. Sahbi. Nonlinear Cross-View Sample Enrichment for Action Recognition. European Conference on Computer Vision. Springer, 2014.
 \bibitem{image_class01}
K. He, X. Zhang, S. Ren, J. Sun. Delving Deep into Rectifiers: Surpassing Human-Level Performance on ImageNet Classification. In IEEE International Conference on Computer Vision (ICCV), 2015

\bibitem{speech_reco01}
A. Graves, A. Mohamed, G. Hinton. Speech recognition with deep recurrent neural networks. In IEEE International Conference on Acoustics, Speech and Signal Processing (ICASSP), 2013
  \bibitem{rbf}
B. Schölkopf, K. Sung, C. Burges, F. Girosi,  P. Niyogi, T. Poggio, V.Vapnik. Comparing support vector machines with gaussian kernels to radial basis function classifiers. A.I.Memo 1599, M.I.T. AI Labs, 1996

\bibitem{lingsahbiicip2014}
L. Wang, H. Sahbi. Bags-of-Daglets for Action Recognition. IEEE International Conference on Image Processing (ICIP), 2014.

\bibitem{polynomial_kernel}
S Amari, S. Wu. Improving support vector machine classifiers by modifying kernel functions. In Neural Net. Vol 12, Issue 6,  783-789, 1999

\bibitem{Mict}
Y. Zhou, X Sun, Z.J Zha, W. Zeng. MiCT: Mixed 3D/2D Convolutional Tube for Human Action Recognition.  In CVPR, 2018
\bibitem{action_localization02}
W. Xu, Z. Miao, J. Yu, Q. Ji. Action recognition and localization with spatial and temporal contexts. Neurocomputing Vol 333, 351-363, 2019

 \bibitem{action_localization03}
H. Zhao, A. Torralba,  L. Torresani, Z. Yan. HACS: Human Action Clips and Segments Dataset
for Recognition and Temporal Localization. In ICCV, 2019
  


\bibitem{action_localization05}
G. Yu, J. Yuan. Fast Action Proposals for Human Action Detection and Search. In CVPR, 2015

\bibitem{action_localization01}
Pointly-Supervised Action Localization. In IJCV, Volume 127, Issue 3, 263–281, 2019

  \bibitem{phong}
P. Vo and H. Sahbi. "Transductive kernel map learning and its application to image annotation." In BMVC,  2012.
\bibitem{speech_reco02}
G.y Hinton, L. Deng, D. Yu, G. Dahl, A. Mohamed, N. Jaitly, A.
Senior, V. Vanhoucke, P. Nguyen, T. Sainath and B. Kingsbury. Deep Neural Networks for Acoustic Modeling in Speech Recognition. In IEEE Signal Processing Magazine, Vol 29: pp. 82-97, 2012
 \bibitem{sahbipr2012}
F. Yuan, G-S. Xia, H. Sahbi, V. Prinet.  Mid-level Features and Spatio-Temporal Context for Activity Recognition. Pattern Recognition.  volume 45, number 12,  4182-4191, 2012
\bibitem{icassp2017b}
M. Jiu and H. Sahbi. Deep kernel map networks for image annotation. IEEE International Conference on Acoustics, Speech and Signal Processing (ICASSP), 2016.

\bibitem{sahbiiccv17}
H. Sahbi. Coarse-to-fine deep kernel networks. Proceedings of the IEEE International Conference on Computer Vision, 1131-1139, 2017. 

\bibitem{image_class02}
T. Xiao, Y. Xu, K. Yang, J. Zhang, Y. Peng, Z. Zhang. The Application of Two-Level Attention Models in Deep Convolutional Neural Network for Fine-Grained Image Classification. In IEEE Conference on Computer Vision and Pattern Recognition (CVPR), 2015


\bibitem{of01}
S. S. Beauchemin, J. L. Barron. The computation of optical flow. ACM Computing Surveys (CSUR) Surveys, Volume 27, Issue 3, Pages 433-466, 1995 
\bibitem{of02}
D. Gu, Z. Wen, W. Cui, R. Wang, F. Jiang, S. Liu. Continuous Bidirectional Optical Flow for Video Frame Sequence Interpolation. In IEEE International Conference on Multimedia and Expo (ICME), 2019
\bibitem{sahbiphd}
H. Sahbi.  Coarse-to-fine support vector machines for hierarchical face detection. PhD thesis, Versailles University, 2003. 

\bibitem{of03}
D. Sun, S. Roth, M. J. Black. Secrets of optical flow estimation and their principles. In IEEE Conference on Computer Vision and Pattern Recognition (CVPR), 2010
\bibitem{of04}
L. Xu, J. Jia, Y. Matsushita. Motion Detail Preserving Optical Flow Estimation. In IEEE Transactions on Pattern Analysis and Machine Intelligence (TPAMI), Volume : 34, Issue : 9, 2012 

  \bibitem{sahbiicassp15}
M. Jiu, H. Sahbi. Semi supervised deep kernel design for image annotation. IEEE International Conference on Acoustics, Speech and Signal Processing, 2015 . 

\bibitem{bag_features}
 G. Csurka, C. R. Dance, L. Fan, J. Willamowski, C. Bray. Visual Categorization with Bags of Keypoints. In European Conference on Computer Vision (ECCV), 2004
   \bibitem{sahbikpca06}
H Sahbi.  Kernel PCA for similarity invariant shape recognition. Neurocomputing 70 (16-18), 3034-3045

 \bibitem{Fisher}
G. Csurka, F. Perronnin. Fisher Vectors : Beyond Bag-of-Visual-Words Image Representations. In International Conference on Computer Vision, Imaging and Computer Graphics, 2010
 \bibitem{pooling01}
K. He, X. Zhang, S. Ren, J. Sun. Spatial Pyramid Pooling in Deep Convolutional Networks for Visual Recognition. In IEEE Transactions on Pattern Analysis and Machine Intelligence (TPAMI), Volume : 37, Issue : 9 , 2015
\bibitem{pooling02}
N. Murray, F. Perronnin. Generalized Max Pooling. In IEEE Conference on Computer Vision and Pattern Recognition (CVPR), 2014
\bibitem{pooling03}
Y. Gao, O. Beijbom, N. Zhang, T. Darrell. Compact Bilinear Pooling. In IEEE Conference on Computer Vision and Pattern Recognition (CVPR), 2016

\bibitem{3dconv01}
D. Tran, L. Bourdev, R. Fergus, L. Torresani, M. Paluri.
Learning spatiotemporal features with 3d convolutional networks. In IEEE International Conference on Computer Vision (ICCV), 2015



\bibitem{dataset_HMDB}
H. Kuehne, H. Jhuang, E. Garrote, T. Poggio, T. Serre.
HMDB: a large video database for human motion recognition. In the International Conference on Computer Vision (ICCV), 2011
\bibitem{pretrained_resnet101_ucf}
J. Yihuang. Pretrained 2D two streams network for action recognition on UCF-101 based on temporal segment network. https://github.com/jeffreyyihuang/two-stream-action-recognition , 2017


\bibitem{VGG}
Wang, Limin, et al. "Places205-vggnet models for scene recognition." arXiv preprint arXiv:1508.01667 (2015).

\bibitem{alexnet}
A. Krizhevsky, I. Sutskever, G.E. Hinton. ImageNet Classification with Deep Convolutional Neural Networks. In Neural Information Processing Systems (NeurIPS), 2012
\bibitem{sahbicvpr08a}
F. Fleuret and H. Sahbi. "Scale-invariance of support vector machines based on the triangular kernel." 3rd International Workshop on Statistical and Computational Theories of Vision. 2003.

\bibitem{MLB-YouTube}
AJ Piergiovanni, M.S. Ryoo. Fine-grained Activity Recognition in Baseball Videos. In IEEE Conference on Computer Vision and Pattern Recognition (CVPR), Workshop on Computer Vision in Sports, 2018
\bibitem{sahbipr19}
H. Sahbi and F.  Fleuret. "Kernel methods and scale invariance using the triangular kernel." (2004).

 

\bibitem{tnnls19}
A. Dutta and H. Sahbi. "High order stochastic graphlet embedding for graph-based pattern recognition." arXiv 1702 (2017)

\bibitem{ntu}
  A. Shahroudy, J. Liu, T. Ng, G. Wang. NTU RGB+D : A Large Scale Dataset for 3D Human Activity Analysis. In IEEE Conference on Computer Vision and Pattern Recognition (CVPR), 2016
\bibitem{ullah2017}
 Ullah, Amin, et al. "Action recognition in video sequences using deep bi-directional LSTM with CNN features." IEEE Access 6 (2017): 1155-1166.
\bibitem{sahbifleuret02}
H. Sahbi  and F.  Fleuret. "Scale-invariance of support vector machines based on the triangular kernel." (2002).

\bibitem{kernels2004}
Shawe-Taylor, John, and Nello Cristianini. Kernel methods for pattern analysis. Cambridge university press, 2004.



\bibitem{icml08}
H. Sahbi,  J-Y.  Audibert, J. Rabarisoa and R. Keriven. "Robust matching and recognition using context-dependent kernels." In Proceedings of the 25th international conference on Machine learning, pp. 856-863. 2008.

\end{thebibliography}
\end{document}